# Active Learning in Brain Tumor Segmentation with Uncertainty Sampling, Annotation Redundancy Restriction, and Data Initialization

**Running Title:** Active Learning in Brain Tumor Imaging

Daniel D Kim[1,2,†], Rajat S Chandra[3,†], Jian Peng[4,†], Jing Wu[4], Xue Feng[5], Michael Atalay[1,2], Chetan Bettegowda[6], Craig Jones[6,7], Haris Sair[6], Wei-hua Liao[4], Chengzhang Zhu[8], Beiji Zou[9], Li Yang[4], Anahita Fathi Kazerooni[10,11], Ali Nabavizadeh[10,12], Harrison X Bai[6], and Zhicheng Jiao[1,2]

[1] Warren Alpert Medical School of Brown University, Providence RI, USA

[2] Department of Diagnostic Imaging, Rhode Island Hospital, Providence, RI, USA

[3] Perelman School of Medicine at the University of Pennsylvania, Philadelphia, PA, USA

[4] Second Xiangya Hospital of Central South University, Changsha, Hunan, China

[5] Biomedical Engineering, University of Virginia, Charlottesville, VA, USA

[6] Department of Radiology, Johns Hopkins University, Baltimore, MD, US

[7] Department of Computer Science, Johns Hopkins University, Baltimore, MD, US

[8] College of Literature and Journalism, Central South University, Changsha, China

[9] School of Computer Science and Engineering, Central South University, Changsha, China

[10] Center for Data-Driven Discovery in Biomedicine (D$^3$b), Children's Hospital of Philadelphia, Philadelphia, PA, USA

[11] Department of Neurosurgery, Perelman School of Medicine, University of Pennsylvania, Philadelphia, PA, USA

[12] Department of Radiology, Perelman School of Medicine, University of Pennsylvania, Philadelphia, PA, USA

[†]DDK, RSC, and JP contributed equally to this work and share co-first authorship

**Corresponding Author:**

Zhicheng Jiao, Department of Radiology, Brown University

593 Eddy St, Rhode Island Hospital, Department of Diagnostic Imaging, Providence, RI, 02903, USA

zhicheng_jiao@brown.edu

**ABSTRACT**

Deep learning models have demonstrated great potential in medical imaging, but their development is limited by the expensive, large volume of annotated data required. Active learning (AL) addresses this by training a model on a subset of the most informative data samples without compromising performance. We compared different AL strategies and propose a framework that minimizes the amount of data needed for state-of-the-art performance. 638 multi-institutional brain tumor MRI images were used to train a 3D U-net model and compare AL strategies. We investigated uncertainty sampling, annotation redundancy restriction, and initial dataset selection techniques. Uncertainty estimation techniques including Bayesian estimation with dropout, bootstrapping, and margins sampling were compared to random query. Strategies to avoid annotation redundancy by removing similar images within the to-be-annotated subset were considered as well. We determined the minimum amount of data necessary to achieve similar performance to the model trained on the full dataset ($\alpha = 0.1$). A variance-based selection strategy using radiomics to identify the initial training dataset is also proposed. Bayesian approximation with dropout at training and testing showed similar results to that of the full data model with less than 20% of the training data ($p$=0.293) compared to random query achieving similar performance at 56.5% of the training data ($p$=0.814). Annotation redundancy restriction techniques achieved state-of-the-art performance at approximately 40%-50% of the training data. Radiomics dataset initialization had higher Dice with initial dataset sizes of 20 and 80 images, but improvements were not significant. In conclusion, we investigated various AL strategies with dropout uncertainty estimation achieving state-of-the-art performance with the least annotated data.

## 1   INTRODUCTION

Deep learning in medical imaging has made significant progress, achieving near or superior performance to that of human expert annotators.[1] Despite the strides in performance, these models are limited by requiring substantial training data and annotations, which are expensive and time-consuming to produce.[2]

Active learning (AL) is a strategy that identifies a subset of unannotated data that would be most informative so that a model can be trained on a subset of annotated samples without compromising performance. Models are built iteratively until acceptable performance is achieved.

AL strategies generally have two approaches: (a) calculating uncertainty and annotating the most uncertain or difficult images or (b) grouping images based on similarity and selecting a subset from each similarity group to identify a representative cohort.[2] To identify uncertain images, Wang et. al uses a preliminary model to predict on unannotated data and assigns images with the smallest probability of the most probable class as most uncertain.[3] An ensemble approach that identifies images with the most disagreement among models can also be used.[4] Bayesian neural networks have alternatively been proposed to use one model to generate a probability distribution instead of a single probability, and wider distributions are attributed to higher uncertainty.[5,6] To reduce annotation redundancy, Yang et. al compares the output from convolutional neural networks, which are ultimately high-level feature vectors, to assess the similarity of unannotated images and identify a representative set of images to annotate.[3] Similarly, traditional computer vision techniques have also been used for feature extraction.[7] Kim et. al combines both uncertainty and representativeness techniques when selecting data to annotate for skin lesion classification and segmentation.[8]

Many AL strategies, including the ones above, focus on 2D imaging, classification tasks, or non-medical imaging.[2-4,6,7] However, application of validated techniques onto 3D medical imaging, such as magnetic resonance imaging (MRI) or computerized tomography (CT), is not straightforward. Some medical imaging tasks have an additional complexity in that they focus on a small region of interest (ROI). Prognosis of brain cancer for example focuses on contrast-enhancing tumor, which is much smaller than the whole brain.[9] This characteristic is exacerbated in 3D imaging as uncertainty calculations need to focus on a small portion of voxels of interest, making them sensitive to noise from the substantial background. Sharma et. al demonstrates remarkable success here by combining least confidence uncertainty estimation and representativeness to create a high performing model using less than 15% of the 2018 Brain Tumor Segmentation (BraTS) dataset.[10] Other works have pursued active learning in 3D medical imaging by incorporating reinforcement learning rather than traditional uncertainty and representative strategies.[11,12]

In this paper, we contribute further to AL in 3D medical imaging with a pilot study by comparing multiple uncertainty and representative techniques and evaluating their individual contribution in reducing the annotation burden on real-world, multi-institutional clinical brain tumor MRI data.

## 2   MATERIALS AND METHODS

### 2.1 Neural Network Architecture

A 5-layered 3D U-net neural network architecture was used.[13] Models used both contrast-enhanced T1-weighted (T1ce) and T2-weighted (T2) sequences to segment the contrast enhancing region. Two patches of size $128 \times 128 \times 128$ biased 25% to the ROI from each image were used for model training. Data augmentation included scaling, rotation, and flipping transformations. Models optimized a soft Dice loss function on the validation set until there was no improvement for 50 epochs. During validation and testing, the full image was inputted. All models were trained using a 16 GB NVIDIA V100 Tensor Core graphical processing unit (GPU). Detailed hyperparameter settings are available in the published study on the full data model.[14]

### 2.2 Active Learning Algorithm

Our proposed methods to improve AL for 3D image segmentation consists of three major components: (1) uncertainty sampling, (2) annotation redundancy restriction, and (3) initial dataset selection (Figure 1).

Uncertainty sampling first uses a model trained on a smaller subset of data to predict segmentations on unannotated data. Uncertainty is calculated from the predicted segmentations, and k images with the highest uncertainty are chosen for annotation. Uncertainty scores estimate the model's confidence on data that was not included in training. Three different uncertainty estimating techniques are outlined below.

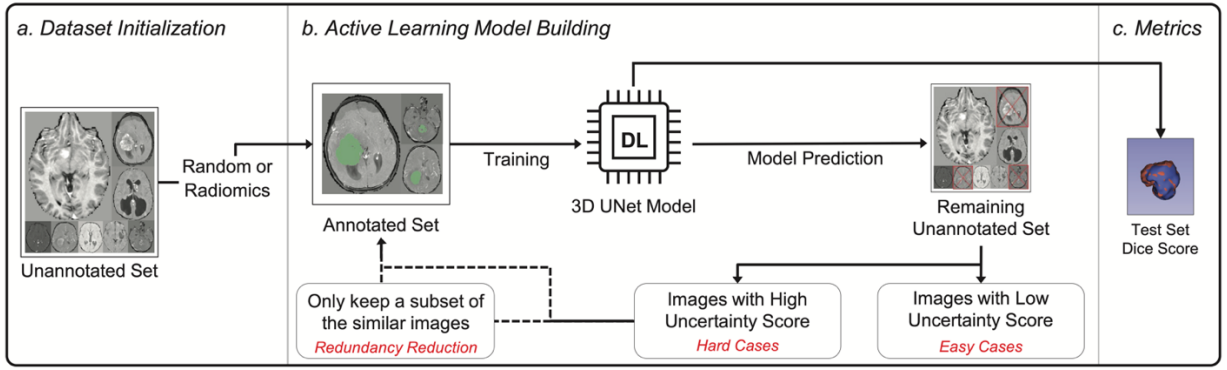

*Figure 1. Workflow of the full active learning framework*

The first technique involves bootstrapping. At each AL iteration, $n$ bootstrapped datasets are generated by sampling with replacement and used to train a separate model. Higher variance in predictions across models suggests ensemble disagreement and higher uncertainty.[4] The uncertainty score is the mean of the variance map of all of the probability maps returned from each bootstrapped model.

Next, we discuss margins sampling. One model is trained on the current batch of annotated data. Each voxel-probability within a probability map $p_i$ for image $i$ can range from values from [0,1]. Voxel-probabilities closer to 0.5 are associated with higher uncertainty.[15] Uncertainty score $u_i$ calculation is described in Equation 1.

$$u_i = -(mean(\,|p_i - 0.5|\,))$$

$$( 1 )$$

Finally, Bayesian models return a probability distribution rather than a single probability. These models can be approximated by generating $n$ predictions from a model that includes a dropout layer.[5] Images with a wider distributions have higher uncertainty. A dropout layer is added to the last decoding convolutional layer in the 3D U-net model. Performance was compared when dropout was enabled during both training and testing versus only at testing. To localize uncertainty estimation to ROIs, the uncertainty score $u_i$ was calculated by taking the mean of the top 0.1% variances of the probability maps, as shown in Equation 2 where $p_{i_j}$ represents the $j$th probability map returned by the model for image $i$.

$$var_i = \mathrm{var}(\mathrm{p_{i_1}}, \cdots, \mathrm{p_{i_n}})$$

$$top_i = \text{highest 0.1\% of values in } \mathrm{var_i}$$

$$\mathrm{u_i} = mean(top_i)$$

$$( 2 )$$

While uncertainty sampling identifies images unfamiliar to the current model, annotation redundancy restriction prevents annotation of images similar to one another. The first annotation redundancy restriction method selects the most representative uncertain images. Consider a subset of $k$ uncertain, unannotated images. If there are $j$ images, where $j < k$, similar to one another, then annotating only some of $j$ images may be sufficient. To evaluate image similarity, we can compare the high-level features between two images. 3D U-net has both an encoder and decoder arm.[13] The encoder arm uses multiple convolutional layers in series to generate an array of high-level features. We modified Yang et al.'s approach of using cosine similarity to compare arrays of high level features for 3D images.[4] The encoder arm returns an 4D array of size (x, y, z, 512), where x, y, z are variable to the size of the input image. The 4D array is then flattened to a 1D array of size 512 by taking the mean across other axes. We can then measure the similarity between two images by calculating the similarity score $ss(I_i, I_j) = cosine\_similarity(h_i, h_j)$ where $h_i$ are flattened high-level features of image $i$ and $h_j$ are that of image $j$. Next, we use the maximum set cover approach to approximate a subset $S_m$ that most represents the entire set of uncertain, unannotated images $S_k$ (Supplementary Algorithm 1).[4] $S_m$ begins as an empty set and is iteratively built to maximize the representativeness of $S_m$ for $S_k$. To quantify how much an image $x \in S_k$ is already represented within $S_m$, we calculate $f(S_m, x) = max_{x_i \in S_m} ss(x_i, x)$. Next, we calculate $F(S_m, S_k) = \sum_{x_i \in S_k} f(S_m, x_i)$ to measure how much $S_m$ represents $S_k$. By iteratively adding the image from $S_k$ that maximizes $F(S_m, S_k)$ until $S_m$ is of size $m$, we can estimate a subset $S_m$ most representative for $S_k$.

The second redundancy restriction method selects uncertain images most non-similar to already annotated data. Consider a subset of $k$ uncertain, unannotated images where there are $j$ images, where $j < k$, similar to the images already annotated. Then, annotating images from $S_j$ may not be informative to the model. In order to get a subset $S_m \subset S_k$ that is most non-similar to the set of already annotated images $S_a$ and images already selected to be annotated, we iteratively build $S_m$ by comparing each image in $S_k$ to the images in $S_a$ and those already in $S_m$, and add the one least similar to these images to $S_m$ from $S_k$ (Supplementary Algorithm 2).

The final investigated AL technique determined the optimal images to initially annotate as opposed to random initialization. We used radiomics feature extraction to build an initial training set with the most diverse features. Radiomic feature extraction was performed with PyRadiomics, an open-source Python package for extracting radiomic features from medical images.[16] The skull-stripped brain region was used as the mask to focus feature extraction. First-order features, including mean, median, entropy, and energy, were extracted to build a feature array per image. Each feature array was normalized across all feature arrays. An initial dataset was then obtained by iteratively selecting the subset of images that maximizes the minimum Euclidean distance between normalized feature arrays similarly to Supplementary Algorithm 1.

2.3 Experiments

We retrospectively collected T2 and T1ce imaging sequences and clinical data from pediatric patients with intracranial leptomeningeal seeding brain tumors who were admitted to 4 large academic hospitals in Hunan Province, China from January 2011 to December 2018 and to the Children's Hospital of Philadelphia (CHOP) from January 2005 to December 2019. Exclusion criteria included patients above 18 years old, patients with missing pathological reports or image sequences, or if images were collected after any tumor-reducing treatment. The institutional review boards of all involved institutions approved this study, and the requirement for informed consent was waived.

Manual segmentation of contrast-enhancing tumor was performed by a neuro-oncologist (J.P.) with 7 years of experience, using the Level Tracing and Threshold tools in 3D Slicer (v.4.10). The MR acquisition parameters are shown in Supplementary Figures 1 and 2. 20% of the data was partitioned as the testing set. At each AL iteration, 20% of the annotated training data was used as the validation set. Images were skull stripped, resampled to isotropic resolution, and co-registered to the same anatomical template. For experimentation purposes, all images were proactively annotated for the contrast-enhancing lesion. However, we assign the images unannotated and annotated states based on the images selected by the AL algorithm. Only images with annotated states were used for model training.

Experiment 1 compared different uncertainty sampling techniques. We initialized a random training set of 40 images and trained a preliminary model. We then annotated 50 of the most uncertain, unannotated images (~10% of entire dataset) returned by the uncertainty sampling technique and retrained the model. Model performance was evaluated at each AL iteration.

Experiment 2 compared different annotation redundancy restriction techniques after uncertainty sampling. We initialized a random training set of 40 images and trained a preliminary model. Of the 100 most uncertain images, we annotated 50 of the most representative images to add to the training data. Because we were interested when the performance of the model is not different to that of the full data model, we use $\alpha = 0.1$ for Experiment 1 and 2.

Experiment 3 compares initial dataset selection using radiomics features of the brain to random initialization. To generate a distribution of model performance trained on random dataset initialization, a model is trained and evaluated on $n$ randomly selected images. This is repeated 10 times. Then, a model is trained on a dataset of size $n$ determined by radiomics and compared in performance to the distribution of random initializations. As we are determining if the methods are statistically different, $\alpha = 0.05$ is used.

## 3    RESULTS

T2 and T1ce sequences with contrast enhancing tumor segmentation were available for 683 brain MRIs from 683 patients (85 Hunan, 598 CHOP). 39 patients were excluded due to skull stripping failure, and 6 were excluded due to co-registration failure. Characteristics for the remaining 638 patients can be found in Table 1. Each model took approximately 1-5 hours to train depending on the AL iteration or size of the annotated training data. The pre-trained models and the AL framework is publicly accessible at https://github.com/naddan27/ActiveLearning/.

Figure 2 compares the mean and median Dice scores of the uncertainty estimation techniques at different percentages of the training data from Experiment 1. Performance of the full data model are taken from its previously published study.[14] Both bootstrapping and Bayesian approximation using dropout at training and testing (dropout traintest) consistently outperformed random query. In contrast, margins and Bayesian approximation using dropout only at testing (dropout test) performed worse than random query. For bootstrapping, there was no significant difference in model performance trained with 27.3% ($p = 0.890$) and 27.3% ($p = 0.874$) of the data versus trained with the full data for mean and median Dice, respectively. For dropout traintest, there was no significant difference in model performance trained with 17.5% ($p = 0.293$) and 17.5% ($p = 0.108$) of the data versus trained with the full data for mean and median Dice, respectively.

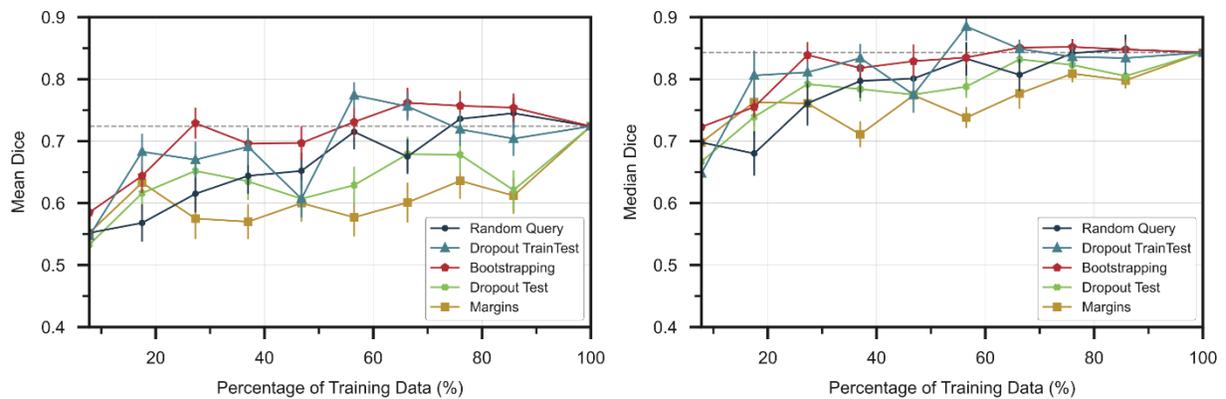

*Figure 2. Dice scores of uncertainty techniques at different percentages of training data. Horizontal dashed line is performance with full training data. Error bars represent the standard error.*

Table 1. Study Population Characteristics

| Characteristics | Number (Percent) |
|---|---|
| Median age at diagnosis in years: Mean (range) | 9.4 (0.1-17.9) |
| Sex | |
|     Male | 355 (55.6%) |
|     Female | 283 (44.4%) |
| Anaplastic astrocytoma | 53 (8.4%) |
| Fibrillary astrocytoma | 35 (5.5%) |
| Glioblastoma | 22 (3.4%) |
| Infiltrating astrocytoma | 20 (3.1%) |
| Pilocytic astrocytoma | 119 (18.7%) |
| Pilomyxoid astrocytoma | 13 (2.1%) |
| Medulloblastoma | 72 (11.3%) |
| Craniopharyngioma | 32 (5%) |
| Dysembryoplastic neuroepithelial tumor （DNET） | 25 (3.9%) |
| Ependymoma | 54 (8.5%) |
| Gangliocytoma/ Ganglioglioma | 58 (9.1%) |
| Meningioma | 22 (3.4%) |
| Neurocytoma | 4 (0.6%) |
| Oligodendroglioma | 8 (1.2%) |
| Pleomorphic xanthoastrocytoma (PXA) | 3 (0.4%) |
| Schwannoma | 4 (0.6%) |
| Subependymal giant cells tumor | 10 (1.6%) |
| Embryonal tumor group | |
|     Atypical teratoid rhabdoid tumor | 18 (2.8%) |
|     Pineoblastoma | 4 (0.6%) |
|     Primitive neuroectodermal tumor | 17 (2.7%) |
| Germ cell tumor group | |
|     Germinoma | 15 (2.4%) |
|     Germ cell tumor | 4 (0.6%) |
| Choroid plexus papilloma | 26 (4.1%) |

In Experiment 2 comparing annotation redundancy restriction techniques, dropout traintest was used to identify the uncertain, unannotated images before annotation redundancy restriction due to its performance in Experiment 1 and smaller computational burden than bootstrapping. Table 2 shows the effect of adding an annotation redundancy restriction technique on top of uncertainty sampling versus just uncertainty sampling alone. While both redundancy restriction techniques achieved similar performance to that of the full model before random query, both redundancy restriction techniques were not able to outperform AL strategies that solely used uncertainty sampling. Figure 3 shows examples of predicted segmentations by models that were built using uncertainty and annotation redundancy restriction AL strategies.

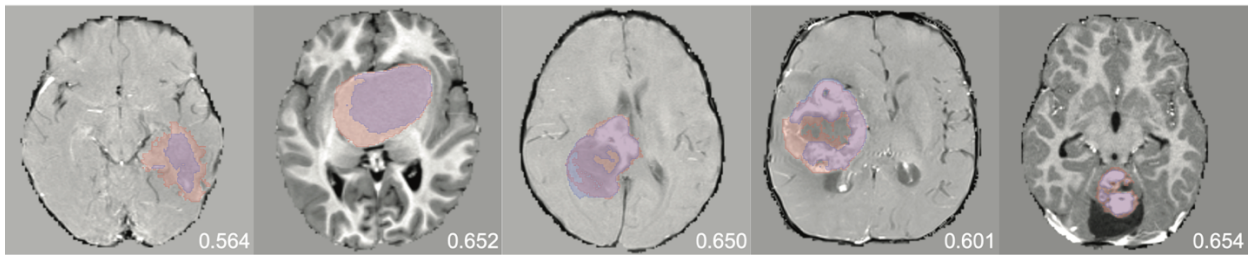

*Figure 3. Examples of predicted (blue) vs expert (red) segmentations at 17.5% of training data with random query, dropout traintest, bootstrapping, redundancy representative, and redundancy non-similar in order. Dice scores are shown for each respective image.*

In Experiment 3 evaluating for a radiomics based initial dataset selection, radiomics dataset initialization had higher Dice at n=20 and 80 than random query, but improvements were not significant. Full comparisons are shown in Table 3.

## 4    DISCUSSION

We compared multiple uncertainty and representative techniques and evaluated their individual and synergistic performance in reducing the number of annotations needed on real-world, multi-institutional clinical data. We show that an AL framework using Bayesian approximation with dropout at training and testing can achieve state-of-the-art performance with less than 20% of the

Table 2. Mean dice of uncertainty and redundancy reduction methods. First iteration without significant difference to performance of full data model are bolded. Standard deviation in parentheses.

| Percentage of Training Data | 7.8% n=40 | 17.5% n=90 | 27.3% n=140 | 37.0% n=190 | 46.8% n=240 | 56.5% n=290 | 66.3% n=340 | 100% n=513 |
|---|---|---|---|---|---|---|---|---|
| Random Query | 0.552 (0.339) | 0.568 (0.331) | 0.615 (0.335) | 0.644 (0.327) | 0.652 (0.326) | **0.715 (0.307)** | 0.675 (0.308) | \| |
| *Uncertainty Only* | | | | | | | | \| |
| Dropout TrainTest | 0.547 (0.329) | **0.683 (0.321)** | 0.670 (0.316) | 0.691 (0.334) | 0.607 (0.345) | 0.774 (0.252) | 0.756 (0.266) | \| |
| Bootstrapping | 0.585 (0.331) | 0.644 (0.319) | **0.729 (0.280)** | 0.696 (0.296) | 0.697 (0.302) | 0.731 (0.280) | 0.762 (0.265) | 0.724 (0.295) |
| *With Dropout TrainTest* | | | | | | | | \| |
| Redundancy Representative | 0.547 (0.329) | 0.561 (0.332) | 0.613 (0.340) | **0.663 (0.315)** | 0.638 (0.315) | 0.612 (0.333) | 0.707 (0.300) | \| |
| Redundancy Non-similar | 0.547 (0.329) | 0.621 (0.336) | 0.590 (0.344) | 0.650 (0.329) | **0.697 (0.302)** | 0.651 (0.328) | 0.730 (0.284) | \| |

*P-values of bolded dice scores: Random Query (p = 0.814), Dropout TrainTest (p = 0.293), Bootstrapping (p = 0.890), Representative (p = 0.115), Non-similar (p = 0.475)*

Table 3. Brain Radiomics Feature Extraction versus Random Selection for Dataset Initialization

| Initial Dataset Size | Random Dice (SD) | Radiomics Dice | p-value |
|---|---|---|---|
| 20 | 0.458 (0.042) | 0.526 | 0.101 |
| 40 | 0.541 (0.030) | 0.513 | 0.349 |
| 80 | 0.535 (0.028) | 0.587 | 0.061 |

training data. Comparatively, AL with only random query achieved full training data model performance at 56.5% of the training data.

We compared 4 different uncertainty estimation techniques to random query: bootstrapping, margins, dropout traintest, and dropout test. While bootstrapping did reduce training data by 70%, its computational demand can be prohibitive. We were therefore interested in using dropout as an alternative. The primary concern with dropout was that it would not be able to generate distinct enough predictions to generate a reliable uncertainty score.[2,6] Furthermore, the prediction variability, which is generally concentrated at the ROI, would be diluted by the substantial number of background voxels in 3D imaging. To address this concern, we presented a dropout strategy that focuses on regions of high disagreement within the image to estimate uncertainty. With this strategy, we show that dropout is generalizable to AL in 3D segmentation tasks and in fact superior to bootstrapping. We also attempted removing dropout training stabilization by implementing dropout only at testing to force more diverse predictions. However, model prediction instability had a stronger negative effect in model performance than the possible positive effects of having diverse predictions for uncertainty estimation as demonstrated by the consistently lower than random query performance even with larger percentages of the training data. Results also show that uncertainty estimation may require calibration given the significant amount of noise contributed by the background voxels in 3D imaging. This may explain margins sampling performing worse than random query in our paper despite other studies showing better performance on 2D imaging.[2]

We were also interested in reducing annotation redundancy. While these techniques were able to achieve similar performance to that of the model trained on the full data with approximately 40%-50% of the training data, they were not able to outperform AL strategies that only incorporate uncertainty. Adding a redundancy restriction strategy can bias training away from uncertain images. Future projects may add a way to prioritize more uncertain images within the representative cohort. Uncertain images can be clustered based on similarity, where each cluster is assigned an overall uncertainty. While only a subset of images from each cluster are selected for annotation, training can be biased toward the uncertain images by artificially increasing images from uncertain clusters with data transformations or generative adversarial networks.[17-21]

Lastly, we evaluate using radiomics features of the brain to construct the initial dataset. At various initial dataset sizes, the performance of the radiomics strategy was not statistically different from random query, demonstrating that selecting an initial dataset based on high level radiomic features of the brain do not translate to selecting images with diverse tumor characteristics. This is understandable given that radiomic features are very dependent on acquisition parameters and heterogeneity in age groups and histologies. An alternative strategy may be to only consider voxels above a specific intensity within the brain mask to vaguely localize the ROI, though this only applies to tasks that segment contrast-enhancing ROI. This AL strategy may be more applicable for models that accept multiple organ systems or imaging modalities to initialize a dataset balanced in all imaging modalities and organ systems of interest.

The major strength of our study is its demonstration of the specific reduction in annotation burden of each AL technique when applied by itself or when combined with multiple techniques. Recent studies have used combined uncertainty and representative approaches,[8,10] but as demonstrated in our uncertainty experiments, AL frameworks are sensitive to calibration when applied to real-world, clinical imaging. By doing a detailed analysis of each technique on annotation burden reduction, our study can guide future studies that combine AL techniques. Furthermore, given the calibration sensitivity and need to optimize hyperparameters at each iteration, our study suggests that AL frameworks may benefit from an adaptative strategy as AL iterations are added. Wang et. al incorporates reinforcement learning with Markov models to create an adaptive AL framework for example,[11] and our study can be used to understand the adaptive strategies returned by reinforcement learning strategies in future studies.

Our study does have limitations. First, we purposefully used the hyperparameters optimized for the full data model on all AL models to address hyperparameter confounding bias, and therefore, results at each iteration may be lower than if they were trained with hyperparameters optimized for each iteration. We assumed performance would be similarly affected for each iteration. Additionally, we only focus on the top 0.1% of variances for the dropout methods when comparing the uncertainty techniques, while the bootstrapping strategy incorporates all variances, making it more sensitive to background voxel noise. Bootstrapping was tested before the incorporation of focusing on the highest variance voxels, and therefore, bootstrapping was not repeated with this technique given that dropout strategy was able to perform similarly to bootstrapping with a significantly lower computational demand. Furthermore, we randomly split the annotated data into the training and validation set at each AL iteration rather than having a consistent validation set at each iteration. Experiments were designed as such with the thought that already deployed AL models should continuously look for more informative samples to add to the training data as time passes and more imaging is available. However, this may bias and overfit models to the training data, hindering model performance at larger AL iterations. Lastly, our generalizability of AL techniques onto real-world, clinical data uses a single dataset and single task of brain tumor segmentation. Further studies with different datasets and tasks are needed before AL strategies can be confidently applied onto real-world data.

## 5    CONCLUSIONS

In conclusion, we assess multiple AL methods and demonstrate their applicability onto real-world clinical 3D brain tumor segmentation. We compare various AL uncertainty estimation as well as annotation redundancy restriction and initial dataset selection strategies, finding that a dropout uncertainty estimation framework is optimal.

### FIGURE LEGEND

**Figure 1.** Workflow of the full active learning framework
**Figure 2.** Dice scores of uncertainty techniques at different percentages of training data. Horizontal dashed line is performance with full training data. Error bars represent the standard error.
**Figure 3.** Examples of predicted (blue) vs expert (red) segmentations at 17.5% of training data with random query, dropout traintest, bootstrapping, redundancy representative, and redundancy non-similar in order. Dice scores are shown for each respective image.

### ACKNOWLEDGEMENTS

Authors declare no acknowledgements. All participating parties gave significant contributions for project development, data availability, annotation, and interpretation and are listed as authors.

### FUNDING

This project was supported by Alpert Medical School Summer Assistantship award to DDK. This work was supported by National Science Foundation of Hunan Province, China (2022JJ30762), International Science and Technology Innovation Joint Base of Machine Vision and Medical Image Processing in Hunan Providence, China (2021CB1013), and the 111 project (B18059) to CZ. This work was supported by Hunan Province Key Areas Research and Development Program, China (2022SK2054) to BZ. This work was supported by Huxiang High-level Talent Gathering

Project-Innovation Talent, China (2021RC5003) to WL. This project was supported by the National Cancer Institute (NCI) of the National Institutes of Health under Award Number R03CA235202 to HXB. This work was supported by the Natural Science Foundation of China (81971696 to LY), Natural Science Foundation of Hunan Province (2022JJ30861 to LY), and Sheng Hua Yu Ying Project of Central South University to LY.

## CONFLICTS OF INTEREST

Authors declare no conflict of interest.

## DATA AVAILABILITY

Data used is not publicly available for patient confidentiality. Interested parties may contact authors regarding questions on data.

**Supplementary Figure 1.** Magnetic field strength and slice thickness of (A) CHOP for T2 and T1-contrast enhanced images

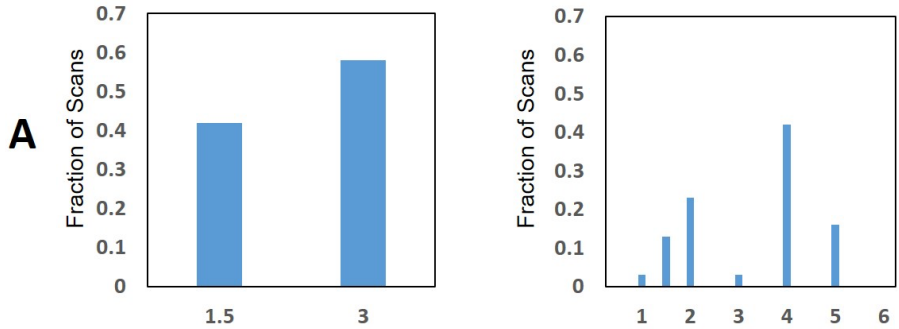

**Supplementary Figure 2.** (A) Echo time for T2 and T1-contrast enhanced MRI images of CHOP; (B) Repetition time for T2 and T1-contrast enhanced MRI images of CHOP

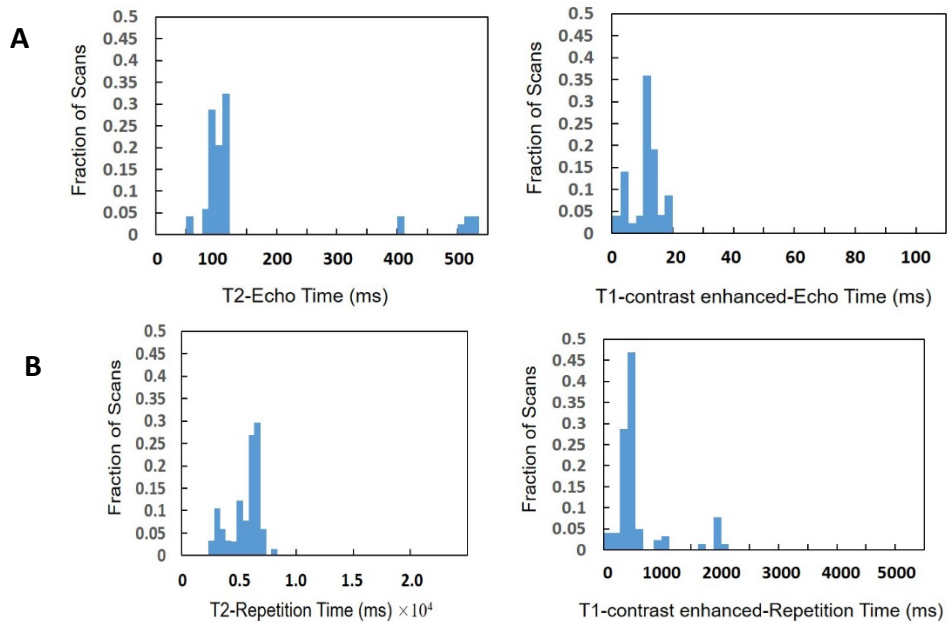

**Supplemental Algorithm 1** Representative subset within uncertain, unannotated images

1: **Initialize** $S_m \leftarrow \{\}$

2: **Initialize** $S_k \leftarrow$ set of $k$ unannotated images with highest uncertainty

3: Move random image from $S_k$ to $S_m$

4: **while** $size(S_m) < m$ **do**

5:     **Set** $F_{max} \leftarrow 0$, $x_{max} \leftarrow none$

6:     **for** all images $x_i \in S_k$ **do**

7:         Temporarily move $x_i$ from $S_k$ to $S_m$

8:          Calculate $F(S_m, S_k)$

9:          **if** $F(S_m, S_k) > F_{max}$ **then**

10:              $F_{max} \leftarrow F(S_m, S_k)$

11:              $x_{max} \leftarrow x_i$

12:          **end if**

13:          Move $x_i$ back from $S_m$ to $S_k$

14:      **end for**

15:      Move $x_{max}$ from $S_k$ to $S_m$

16: **end while**

---

**Supplemental Algorithm 2** Subset of uncertain unannotated images non-similar to already annotated images

---

1: **Initialize** $S_m \leftarrow \{\}$

2: **Initialize** $S_k \leftarrow$ set of $k$ unannotated images with highest uncertainty

3: Move random image from $S_k$ to $S_m$

4: **while** $size(S_m) < m$ **do**

5:      **Set** $ss\_sum_{min} \leftarrow +\infty$, $x_{min} \leftarrow none$

6:      **for** all images $x_i \in S_k$ **do**

7:          $similarity\_sum \leftarrow 0$

8:          **for** all images $x_j \in (S_a \cup S_m)$ **do**

9:              $similarity\_sum \mathrel{+}= ss(x_i, x_j)$

10:         **end for**

11:         **if** $similarity\_sum < ss\_sum_{min}$ **then**

12:             $ss\_sum_{min} \leftarrow similarity\_sum$

13:             $x_{min} \leftarrow x_i$

14:         **end if**

14:     **end for**

15:     Move $x_{min}$ from $S_k$ to $S_m$

16: **end while**